\documentclass{article}

\usepackage{microtype}
\usepackage{graphicx}
\usepackage{amsmath,amssymb}
\usepackage{bm}
\usepackage{multirow}
\usepackage{booktabs} % for professional tables
\usepackage{caption}
\usepackage{subcaption}
\usepackage{xcolor}

\usepackage{hyperref}

\usepackage{setspace}

\usepackage[accepted]{icml2021}

\icmltitlerunning{Ordinal Regression via Binary Preference vs Simple Regression: Statistical and Experimental Perspectives}
\begin{document}

\twocolumn[
\icmltitle{Ordinal Regression via Binary Preference vs Simple Regression:\\ Statistical and Experimental Perspectives}



\icmlsetsymbol{equal}{*}

\begin{icmlauthorlist}
\icmlauthor{Bin Su*}{ms,thu}
\icmlauthor{Shaoguang Mao}{ms}
\icmlauthor{Frank Soong}{ms}
\icmlauthor{Zhiyong Wu}{thu}
% \icmlauthor{Fiuea Rrrr}{to}
% \icmlauthor{Tateu H.~Yasehe}{ed,to,goo}
% \icmlauthor{Aaoeu Iasoh}{goo}
% \icmlauthor{Buiui Eueu}{ed}
% \icmlauthor{Aeuia Zzzz}{ed}
% \icmlauthor{Bieea C.~Yyyy}{to,goo}
% \icmlauthor{Teoau Xxxx}{ed}
% \icmlauthor{Eee Pppp}{ed}
\end{icmlauthorlist}

\icmlaffiliation{ms}{ Microsoft Research Asia}
\icmlaffiliation{thu}{ Tsinghua-CUHK Joint Research Center for Media Sciences, Technologies, and Systems, Shenzhen International Graduate School, Tsinghua University}
% \icmlaffiliation{ed}{School of Computation, University of Edenborrow, Edenborrow, United Kingdom}

\icmlcorrespondingauthor{Shaoguang Mao}{shaoguang.mao@microsoft.com}
% \icmlcorrespondingauthor{Eee Pppp}{ep@eden.co.uk}

\icmlkeywords{Machine Learning, ICML}

\vskip 0.3in
]



\printAffiliationsAndNotice{\icmlEqualContribution}  % leave blank if no need to mention equal contribution

\begin{abstract}
% This document provides a basic paper template and submission guidelines.
% Abstracts must be a single paragraph, ideally between 4--6 sentences long.
% Gross violations will trigger corrections at the camera-ready phase.

% Ordinal regression is a special regression where the labels of the target variable exhibit a nature ordering. There's a some ordinal regression methods have shown its improvement on a lot of tasks, however, most proposed ordianl regression methods based on human perceptual cognition and are verified by experiments. When a method is used in other tasks or fields, it is impossible to know whether it is effective or not except through experiments. This article analyzes the possible reasons why a special ordinal regression with achored reference samples (ORARS) performs better than the regression method, and shows that this method can achieve better results on general regression problems. 

Ordinal regression with anchored reference samples (ORARS) has been proposed for predicting the subjective Mean Opinion Score (MOS) of input stimuli automatically. The ORARS addresses the MOS prediction problem by pairing a test sample with each of the pre-scored anchored reference samples. A trained binary classifier is then used to predict which sample, test or anchor, is better statistically. Posteriors of the binary preference decision are then used to predict the MOS of the test sample. In this paper, rigorous framework, analysis, and experiments to demonstrate that ORARS are advantageous over simple regressions are presented. The contributions of this work are: 1) Show that traditional regression can be reformulated into multiple preference tests to yield a better performance, which is confirmed with simulations experimentally; 2) Generalize ORARS to other regression problems and verify its effectiveness; 3) Provide some prerequisite conditions which can insure proper application of ORARS.

\end{abstract}

\section{Introduction}
Ordinal regression is a special regression task where the labels of the target variable exhibit a nature ordering \cite{7161338}. As \cite{odd} suggested, the ordinal regression mainly includes two categories, i.e., ``grouped continuous variables" and ``assessed ordered categorical variables". The first one is to predict the discretized continuous variable and the second one is to predict categorical human judgements which is represented by given labels, e.g. poor, good, excellent, etc.

In this kind of tasks, the input samples in datasets are with natural ordering. The use of ordinal information can improve the prediction results, which is confirmed in a lot of domain, such as medical research \cite{medcial1,medcial2}, credit rating \cite{credit1,credit2} and etc. 

Ordinal regression is a task of both classification and regression, and there's a lot of methods developed to solve this problem. A na\'ive approach is solve it via either a traditional classification \cite{Diaz_2019_CVPR} or a regression approach \cite{torra2006regression}. However, in these methods, the ordinal or the ranking information is not utilized, at least not explicitly.

\cite{lin2012reduction} solved the ordinal regression with binary decomposition by deciding whether a test sample is better than a preset grade. \cite{perez2013projection} presented an improved approach, which solved the ordinal regression task with a three decisions, i.e., lower, equal, or higher. Decomposing the ordinal regression into a series of classifications is a classical ordinal regression solution, and some similar approaches have been tried, e.g. \cite{mccullagh1980regression,herbrich1999support,xia2007ordinal,cheng2008neural}. 

These methods utilized the ordinal information between a given sample and a given ranks. In fact, ordinal information is a partial order relationship, and broadly exists between any sample pair. By comparing multiple paired samples, the ordinal or rank order can be better learned. 

\cite{mao2019nn} proposed a new kind of ordinal regression method, called Ordinal Regression with Anchored Reference Samples (ORARS), which picked some pre-scored samples as reference. The new method is inspired by that the AB test is easier than mean opinion score (MOS) for human subjects, and thus build a binary classifier to judge which one is better given two samples instead of building models for each rank. The final score is calculated by comparing test sample with all pre-scored samples in each rank.
\begin{figure*}[!t]
    \centering
    \includegraphics[width=14cm]{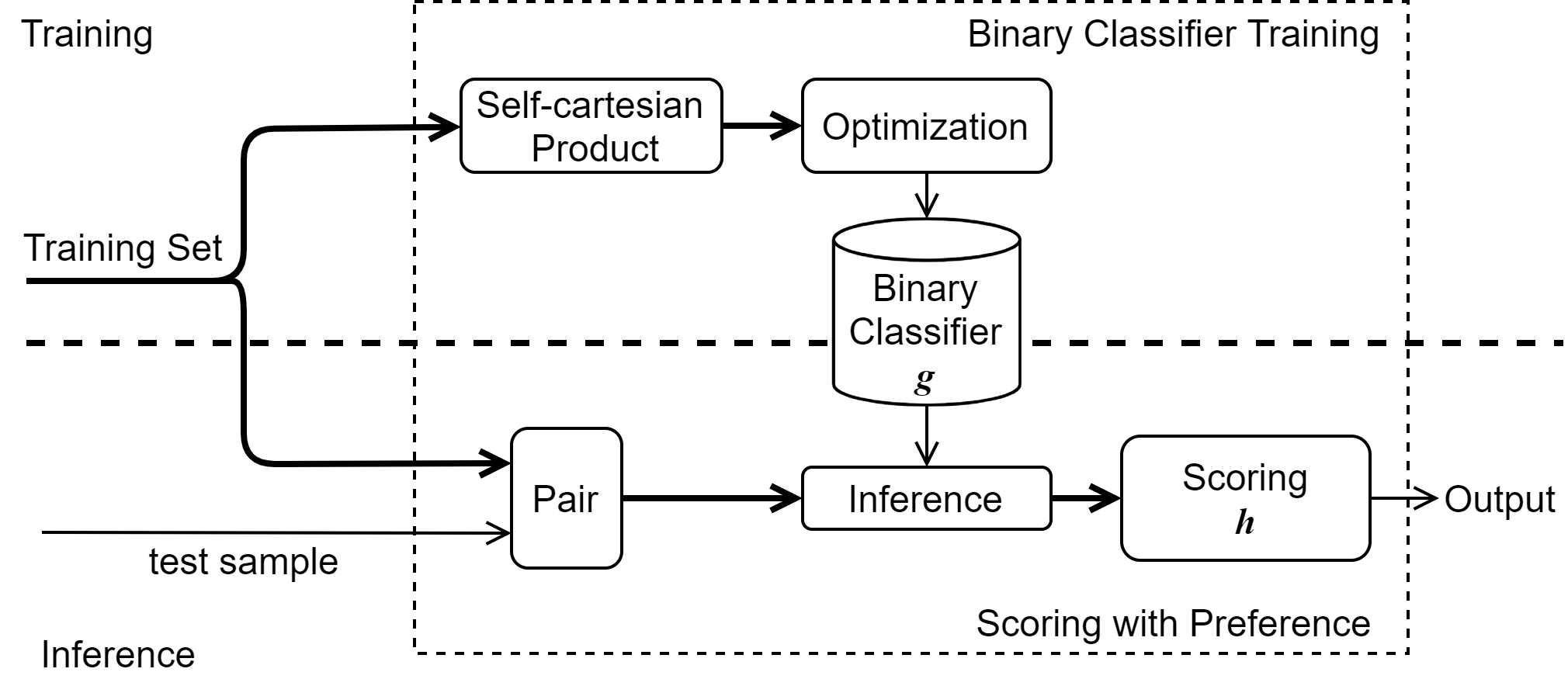}
    \caption{Framework of Ordinal Regression with Anchored Reference Samples (ORARS)}
    \label{fig:framework}
\end{figure*}

However, one flaw with the existing approaches, including \cite{mao2019nn}, is that they can only handle discrete labels. For example, the prediction target must be like \{1,2,3,4,5\} in a five-marked assessment. It means these methods cannot be applied in a continuous regression task.

\cite{su2020improving} improved the scoring method in ORARS to make it possible to apply ORARS to imbalanced datasets. The new scoring method is used to assess the test sample with the score distribution of the pre-scored reference samples. Because the proposed model is not associated with specific label, it can be applied to continuous regression tasks.

With previous experimental observations on ORARS's benefits, this paper is interested in whether the ORARS could improve the performance against traditional regression and expands it to more application scenarios.

The rest of the paper is organized as follows: In section two, we formalize ORARS and investigate how each component of ORARS can yield a better performance with mathematical derivation and experimental simulations. In section three, experimental results on regression benchmark datasets to show the effectiveness are presented. In section four, more insights on ORARS are presented. In the last section, a conclusion is drawn.

\section{Ordinal Regression with Anchored Reference Samples}

Riding on the works of \cite{mao2019nn} and \cite{su2020improving}, the ORARS framework for continuous variable regression is shown as Figure.\ref{fig:framework}. The ORARS system consists of two major components: \textbf{Binary Classifier Training} and \textbf{Scoring with Preference}. In \textbf{Binary Classifier Training}, a binary classifier is trained to determine which sample in formed sample pair performs better with their corresponding features. 
In \textbf{Scoring with Preference}, the score of a test sample is predicted by comparing the test sample with all anchored reference samples.

\subsection{Formalization}
\label{sec:form}

To better explain and analyze ORARS, we formalize it as follows.

\paragraph{Binary Classifier Training}
Let $\bm{x}\in\mathbb{R}^n$ and $y\in\mathbb{R}$ be the feature vector and its corresponding ground-truth score, respectively, where $\mathbb{R}$ and $\mathbb{R}^n$ indicate the real number field and the $n$, the feature dimension, real number space correspondingly. The binary classifier,  $g:\mathbb{R}^{2n}\rightarrow\mathbb{R}$, uses the concatenation of the test sample's feature $\bm{x_t}$ and an anchored reference sample's feature $\bm{x_a}$ as input and computes the posterior probabilities $p\in[0,1]$, denoted confidence that the test sample is better than the given anchored reference sample. The process could be written as 
\begin{equation}
    p=g(\bm{x_t},\bm{x_a})
    \label{eq:g}
\end{equation}
A vectorized version would be:
\begin{equation}
    \bm{p}=g(\bm{x_t},\mathbf{X_A})
    \label{eq:vec_g}
\end{equation}
where the test sample is compared with all anchored reference samples. The $\mathbf{X_A}$ denotes all sample features in the anchor set $\mathbf{A}$ and $\bm{p}$ is a vector of all posteriors in Equation.\ref{eq:g}.

In particular, with a training dataset $\mathbf{D}$, to train the binary classifier, $g$, training pair set $\mathbf{P}$ is generated with the self-cartesian product on $\mathbf{D}$. Let ${(x_i,y_i)}$ and ${(x_j,y_j)}\in\mathbf{D}$ be the $i$-th and $j$-th sample, respectively. The label $t_{ij}$ generated from this sample pair is shown Equation.\ref{eq:get_label}
\begin{equation}
    t_{ij}=\left\{
    \begin{aligned}
    1\quad& y_i>y_j \\
    0\quad& y_i\le y_j 
    \end{aligned}
    \right.
    \label{eq:get_label}
\end{equation}
In this paper, all samples in $\mathbf{D}$ are employed as anchored reference samples, i.e. $\mathbf{D}$ is $\mathbf{A}$. 

\paragraph{Scoring with Preference}
The scoring process takes computed posteriors of the test sample paired with all anchored reference samples as input. The scoring function is formulated as $h:\mathbb{R}^{|\mathbf{A}|}\rightarrow\mathbb{R}$, where $|\mathbf{A}|$ is the size of anchor set $\mathbf{A}$. In this paper, a special case of $h$ is adopted as scoring methods:
\begin{equation}
    h(\bm{p},\bm{y_a})=\bm{y_a}'[max(|\mathbf{A}|-1,\lfloor\sum_{p\in\bm{p}}p\rfloor)]
    \label{eq:score_alg}
\end{equation}
where $\bm{p}$ is the vector in Equation.\ref{eq:vec_g},  $\bm{y_a}'$ is the sorted score sequence of all anchored reference samples in ascending order, $\lfloor\ \rfloor$ is a round-down operation and $\bm{y}[k]$ indicates outputting the $k$-th element (index starts from 0) in the sequence $\bm{y}$. The $max$ operation is used to avoid out of bound error.

Generally speaking, the test sample is compared with all samples in the training set. Then its relative order among training set is predicted. Finally, the corresponding score are outputted as prediction.

\subsection{Theoretical Analysis}
For a continuous variable regression task, traditional regression models map the input feature vector to a score directly, i.e. constructing $f:\mathbb{R}^n\rightarrow\mathbb{R}$ to map an N-dimensional input to a score. To compare ORARS with simple regression, an intermediate framework, called simplified ORARS (sORARS), is introduced.

The sORARS is modified from ORARS with a rule-based binary decision $g$. Let function $f$ be a trained regression model. A simplified $g$ in sORARS is shown as Equation.\ref{eq:spec_g}.
\begin{equation}
    g(\bm{x_t},\bm{x_a})=\left\{
    \begin{array}{ll}
        1 & f(\bm{x_t})>f(\bm{x_a})\\
        0 & f(\bm{x_t})\le f(\bm{x_a})\\
    \end{array}
    \right.
    \label{eq:spec_g}
\end{equation}
With the introduction of this simplified $g$, the sORARS adopts the scoring module $h$ to equalize the distribution of regressor $f$'s outputs with the distribution of the anchor set. When sORARS contrasts with traditional regression, the benefits from \textbf{Scoring with Preference} $h$ will be observed and when sORARS contrasts with ORARS, the benefits from \textbf{Binary Classifier Training} can be observed.

\subsubsection{Regression vs sORARS}
\label{sec:reg_sorars}
%To observe the gain from the \textbf{Scoring with Preference} module, simulations under given assumptions and experiments in real conditions are conducted.

For sORARS, a series of comparisons between a test sample and all anchored reference samples are conducted with the assistance of a trained regressor $f$. By imposing some assumptions on the output of the regressor $f$, analysis of whether the \textbf{Scoring with Preference} module can improve the output of $f$ can be conducted mathematically.
%Since the parameter part of sORARS consists of only one regressor which can share trained regression model. In the comparative analysis of regression and sORARS, only the output of the regressor is important rather than how to get the output of the regressor. 

%By imposing some strict restrictions on the output of the regressor $f$, analysis of whether sORARS can improve the output of the regressor can be conducted in mathematical persepective. 

In this chapter, two sets of theoretical analysis to show gains from $h$ are included. First, a proof is given with the assumption that the distribution of the regressor prediction error is under a uniform distribution on all scores. Meanwhile, a corresponding simulation is presented to echo the proof. Second, under a more complicated but realistic condition where the regressor prediction error follows a zero-mean normal distribution on all score, mathematical derivation is challenging, so simulation is presented for the analysis.

\paragraph{Uniform Distribution of Error}
The random variable $S$ denotes the prediction result from the trained regressor $f$ and the random variable $Y$ denotes the ground truth. Assume that 
\begin{equation}
    Y\sim U(0,N)
    \label{eq:y_dist}
\end{equation}
which means $Y$ obeys a uniform distribution between 0 and N. 

The restriction that the distribution of regressor prediction error follows a uniform distribution on all score is assumed as in Equation.\ref{eq:sy_dist}
\begin{equation}
    S-Y\sim U(-M,M)
    \label{eq:sy_dist}
\end{equation}
Thus, the Mean Absolute Error (MAE) between $S$ and the ground-truth $Y$ could be computed as in Equation.\ref{eq:regressor_mae}
\begin{align}
    \mathbb{E}|S-Y|& =\int_{-M}^{M}\frac{1}{2M}|x|dx\\
    &=\frac{M}{2}
    \label{eq:regressor_mae}
\end{align}

The MAE of sORARS is derived from the error rate of binary preference decision, $\xi$, which is determined by the prediction errors from the regressor $f$. 
%By applying the scoring module $h$, the MAE of sORARS could be calculated with a given error rate $\xi$ in a statistical perspective.

To derive the error rate of binary preference decision, $\xi$ of $g$ described in Equation.\ref{eq:spec_g}, two intermediate random variables are defined in Equation.\ref{eq:z} and Equation.\ref{eq:e} 
\begin{align}
    Z\equiv Y_1-Y_2 \label{eq:z} \\
    E\equiv S_1-S_2 \label{eq:e}
\end{align}

where 
\begin{align}
    Y_1 &\sim U(0,N)\\
    Y_2 &\sim U(0,N)\\
    S_1 &\sim U(-M,M)\\
    S_2 &\sim U(-M,M)
\end{align}
When $Z<0$, the Cumulative Density Function (CDF) of $Z$ could be written as Equation.\ref{eq:cdf_z}
\begin{align}
    F(z)&=\int_{0}^{N+z}\int_{y_1-z}^{N}\frac{1}{N^2}d{y_2}d{y_1}\\
    % &=\frac{1}{N^2}\int_{0}^{N+z}(N+z-x_1)d{x_1}\\
    &=\frac{(N+z)^2}{2N^2}\label{eq:cdf_z}
\end{align}
and corresponding Probability Density Function (PDF) is as in Equation.\ref{eq:pdf_z}
\begin{equation}
    p(z)=\frac{N+z}{N^2}\label{eq:pdf_z}
\end{equation}

Similarly, when $E<0$, the PDF of $E$ is calculated as in Equation.\ref{eq:pdf_e}
\begin{equation}
    p(e)=\frac{2M+e}{4M^2}\label{eq:pdf_e}
\end{equation}

% \begin{figure}
%     \centering
%     \includegraphics[width=8cm]{fig/uniform.png}
%     \caption{Simulation result with \textbf{uniform distribution of error}. The horizontal axis is the number of samples in one simulation. The vertical axis indicates the upper bound of the regressor prediction error. Each point represents the MAE improvement of a simplified ORARS relative to a simulated traditional regression method under a 100-point scale assessment, with a given sampling.}
%     \label{fig:uniform_sim}
% \end{figure}

The error rate $\xi$ could be written as in Equation.\ref{eq:orerr}
\begin{align}
    \xi&=2P(e<z,e<0,z<0)\\
    % &=2\frac{1}{4M^2N^2}\int_0^{2M}(N-z)\int_{z}^{2M}(2M-e)dedz\\
    % &=\frac{1}{4M^2N^2}\int_0^{2M}(N-z)(2M-z)^2dz\\
    &=\frac{2NM-M^2}{3N^2}\label{eq:orerr}
\end{align}

After the scoring processing $h$, the MAE of sORARS could be written as Equation.\ref{eq:mae_sorars}
\begin{align}
    % \mathrm{MAE}&=\int_0^N|y-y_t|dy_t\\
    \mathrm{MAE_{sORARS}}&=\int_0^N|\frac{(1-\xi)y+\xi(N-y)}{N}N-y|dy\\
    % &=\int_0^N|\xi(N-2y_t)|dy_t\\
    % &=2\xi\int_0^{\frac{N}{2}}(N-2y_t)\frac{1}{N}dy_t\\
    % &=\frac{2\xi}{N}*\frac{N^2}{4}\\
    &=\frac{\xi N}{2} \label{eq:mae_xi}
    % &\le \frac{M}{3} \label{eq:mae_sorars}
\end{align}
Replace $\xi$ with Equation.\ref{eq:orerr}, the $\mathrm{MAE_{sORARS}}$ is presented as Equation.\ref{eq:mae_sorars}.
\begin{align}
    \mathrm{MAE_{sORARS}}&=\frac{M}{3}-\frac{M^2}{N}\\
    &\le \frac{M}{3} \label{eq:mae_sorars}
\end{align}

The contrast between Equation.\ref{eq:regressor_mae} with Equation.\ref{eq:mae_sorars} shows that the sORARS proposal absolutely reduce MAE with at least $\frac{M}{6}$ and
achieves a better performance under this condition. 

A series of simulations are conducted to confirm the proof. In the simulation, the $N$, i.e. scoring scale, is set to $100$. The first step is to sample $C$ points from $U(0,N)$ to form the ground truth vector $\bm{y}$, then $C$ error values are sampled from $U(-M,M)$ to form the error vector $\bm{e}$. Hence, the output of corresponding regressor is
\begin{equation}
    \bm{s}=\bm{y}+\bm{e}
    \label{eq:recover_s}
\end{equation}

The MAE of the fictitious regressor could be computed as 
\begin{equation}
    \mathrm{MAE_{reg}}=\frac{1}{C}\sum_{e\in\bm{e}}|e|
    \label{eq:sim_reg_mae}
\end{equation}

With all other $C-1$ points as anchored references, each point could be predicted as a new score $s'$ with Equation.\ref{eq:score_alg}, the MAE between new scores and $\bm{y}$ could be $\mathrm{MAE_{sORARS}}$.
%\begin{equation}
%    s'=sORARS(s,\bm{s}-,\bm{y})
%\end{equation}
The $\mathrm{MAE_{gain}}$ is 
\begin{equation}
    \mathrm{MAE_{gain}}=\mathrm{MAE_{reg}}-\mathrm{MAE_{sORARS}}
    \label{eq:mae_gain}
\end{equation} 
$\mathrm{MAE_{gain}}$ is the relative improvement or degradation of sORARS against regression: when sORARS performs better than the regression, it is positive and vice versa.

\begin{figure*}
     \centering
     \begin{subfigure}[t]{0.48\textwidth}
         \centering
        \includegraphics[width=\textwidth]{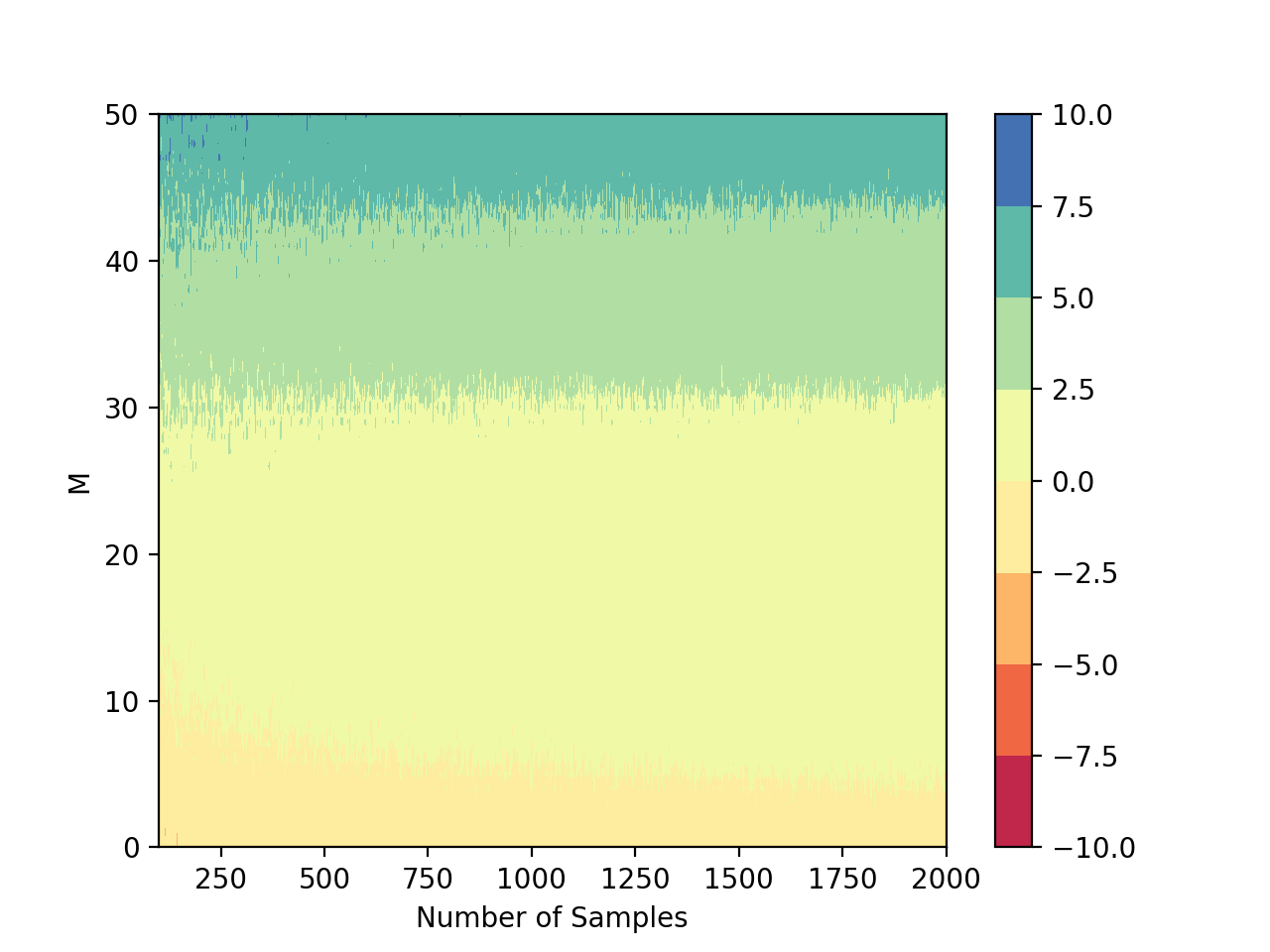}
        \caption{\textbf{Uniform distribution of error}. The horizontal axis is the sampling number in one simulation. The vertical axis indicates the upper bound of the regressor prediction error. Each point represents the MAE improvement of a simplified ORARS relative to a simulated traditional regression method under a 100-point scale assessment, with a given sampling.}
        \label{fig:uniform_sim}
     \end{subfigure}
     \hfill
     \begin{subfigure}[t]{0.48\textwidth}
         \centering
         \includegraphics[width=\textwidth]{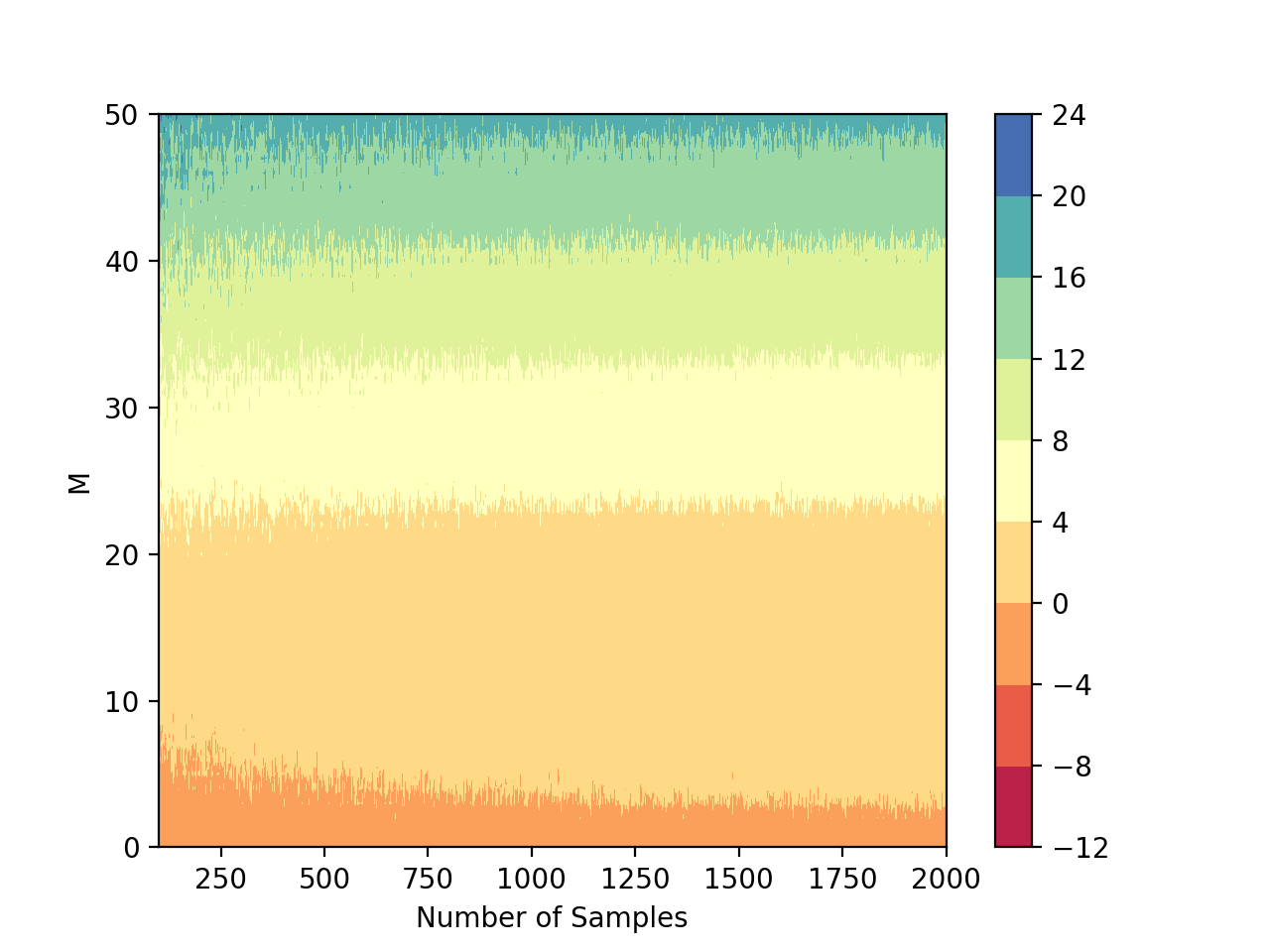}
         \caption{\textbf{Normal distribution of error}. The horizontal axis is the sampling number in one simulation. The vertical axis indicates the standard deviation of the regressor prediction error. Each point represents the MAE improvement of a simplified ORARS relative to a simulated traditional regression method under a 100-point scale assessment, with a given sampling.}
         \label{fig:normal_sim}
     \end{subfigure}
     \hfill
    \caption{Simulation results}
    \label{fig:vis}
\end{figure*}

Figure.\ref{fig:uniform_sim} shows the change of $\mathrm{MAE_{gain}}$ with the number of sampling points and the error hyper-parameter $M$ in simulation. Two conclusions can be drawn: 1) when samples in one simulation is adequate (more than 500) and the original regression's largest bias is larger than a threshold (5 in 100-point scale), the sORARS gain is stable. 2) take specific cases ($M$=50, $C>$ 500) as examples, their corresponding $\mathrm{MAE_{gain}}$ is between 7.5 to 10, which align with previous mathematical proof that an $\frac{M}{6}$ improvement will be obtained.

% \begin{figure}
%     \centering
%     \includegraphics[width=8cm]{fig/normal.png}
%     \caption{Simulation result with \textbf{normal distribution of error}. }
%     \label{fig:normal_sim}
% \end{figure}

\paragraph{Normal Distribution of Error} A more realistic assumption of the regressor prediction error is a zero-mean normal distribution. $S-Y\sim N(0,M)$ where $N(0,M)$ is a normal distribution with 0 and M as the mean and standard deviation respectively.

The simulation is conducted with the same procedure in uniformly distributed error case and the corresponding result is shown as Figure.\ref{fig:normal_sim}. As the simulation shows, when the sampling number in one simulation is adequate (more than 500) and the standard deviation of regressor prediction error is beyond a threshold ($M>3$, which is a tiny value in a 100-point scale assessment), the sORARS can improve MAE stably. 

In a word, with a mathematical proof and corresponding simulation evidence, we can draw the conclusion that the \textbf{scoring with preference} strategy $h$, which is the key component in ORARS, can indeed improve the performance against a simple regression. 

Intuitively speaking, the scoring with preference strategy uses all samples in training set as reference, which constructs a more real data distribution, so the predictions are closer to the real data distribution than traditional regression.

%\begin{figure}
%    \centering
%    \includegraphics[width=8cm]{fig/normal.png}
%    \caption{Simulation result with \textbf{normal distributed error}.  Thehorizontal axis is the size of simluation dataset, the vertical axis is the standard deviation of the normal distribution. Each point represents the MAE improvement of a special ORARS relative to the traditional regression method under a 100-point scale assessment, a given sample size and a given error.}
%    \label{fig:normal_sim}
%\end{figure}

% Figure.\ref{fig:realcase} showed how the ORARS improve a real regressor's output where the error distribution of linear regressor is shown as Figure.\ref{fig:real_error}.

% \begin{figure}
%     \centering
%     \includegraphics[width=6cm]{fig/linearRegressionDist.png}
%     \caption{Caption}
%     \label{fig:realcase}
% \end{figure}
% \begin{figure}
%     \centering
%     \includegraphics[width=6cm]{fig/linearRegressionError.png}
%     \caption{Caption}
%     \label{fig:real_error}
% \end{figure}

\subsubsection{ORARS vs sORARS}
As mentioned in Equation.\ref{eq:spec_g}, a special case of $g$ is introduced for analysis. From an optimization perspective, it is intuitively and empirically correct that by optimizing the binary classifier directly, a more reliable model could be obtained compared with modeling a regressor $f$ first and making binary preference decisions based on a specific rule.

In particular, the binary classifier can be trained with different machine learning methods, including neural network models. So here the discussions on why a direct optimization outperforms regression plus a specific rule are omitted.  Considering the only difference between ORARS and sORARS is the binary decision $g$, we provide an verification from experimental perspective. A series of experiments on benchmark datasets are conducted with ORARS and sORARS to demonstrate that it is easy to find a better $g$ with 
supervised classification methods. More details will be elaborated in the \textbf{Experiments} session.

\section{Experiments}
Experiments in \cite{mao2019nn,su2020improving} confirmed the superiority of ORARS in two typical ordinal regression tasks, i.e. speech fluency assessment and speech pronunciation evaluation. Actually, ORARS is not only suitable for ordinal regression but for continuous variable regression. The following experiments are conducted on continuous regression tasks to show the expanded significance of ORARS in regression tasks.
%the ORARS is not only suitable for ordinal regression task but also is applicable for continuous variable regression tasks for the analysis without introdcing any nature of ordinal regression.

\subsection{Dataset}

The experiments are conducted on the regression benchmark datasets provided by \cite{chu2005gaussian} in sixteen regression tasks. Both continuous version and discretized versions are provided. The continuous version is adopted and its detailed characteristics are shown in  Table.\ref{tab:benchmark_result}. The name of each dataset is consistent with the names mentioned in \cite{chu2005gaussian}.

As \cite{chu2005gaussian} proposed, the experiments on each dataset are conducted with twenty-fold cross-validation. The data is divided for traditional machine learning experiments, where each model is trained on a smaller dataset and tested on a larger dataset. The division of the dataset is not accepted for experiments on deep-learning based model training. In this paper, all experiments are conducted with a five-fold cross-validation and in each fold, the whole dataset is splitted into 80\% and 20\% as training and testing sets correspondingly. In the training set, 10\% is reserved as development set and the rest is used as the true training set.
% \footnote{Detailed information on data split can be found in 	supplementary materials for reproduction.}
%To be compared with traditional machine learning methods, some representative methods reported in \cite{gutierrez2015ordinal} are reproduced and tested on the new data splitting. 
% Another difference is the label will not be discretized for the proposed method no discretization restriction.

% \begin{table*}[]
%     \centering
%     \begin{tabular}{c|c|c|c|c}
%     \hline
%         Name & Size & Min & Max & Attributes\\ \hline
%         Diabetes & 43 & 3.0 & 6.6 & 2\\ 
%         Pyrimidines & 74 & 0.44 & 0.9 & 27\\
%         Triazines & 186& 0.1& 0.9 & 60\\
%         Wisconsin Breast Cancer & 194& 1.0 & 125.0 & 32 \\
%         Machine CPU & 209 &6.0& 500 & 6\\
%         Auto MPG & 392 &9.0 &46.6 & 7\\
%         Boston Housing & 506 & 5.0 & 50.0& 13 \\
%         Stocks Domain & 950 & 34.0 & 62.0 & 9\\
%         Abalone & 4177 & 1.0 & 29.0 & 8\\
%         Bank Domains(1) & 8192& 0.00& 0.80&8\\
%         Bank Domains(2) & 8192& 0.00 &0.820& 32\\
%         Computer Activity(1) &8193&0.0 &99.0 & 12\\
%         Computer Activity(2) &8193&0.0 &99.0 & 21\\
%         California Housing & 20640& 14999 & 500001 &8\\
%         Census Domains(1)  & 22784& 0& 500001 &8\\
%         Census Domains(2) & 22784& 0& 500001 & 16\\ \hline
%     \end{tabular}
%     \caption{Caption}
%     \label{tab:benchmark_info}
% \end{table*}

\subsection{Experimental Settings}
\begin{figure}
    \centering
    \includegraphics[width=7cm]{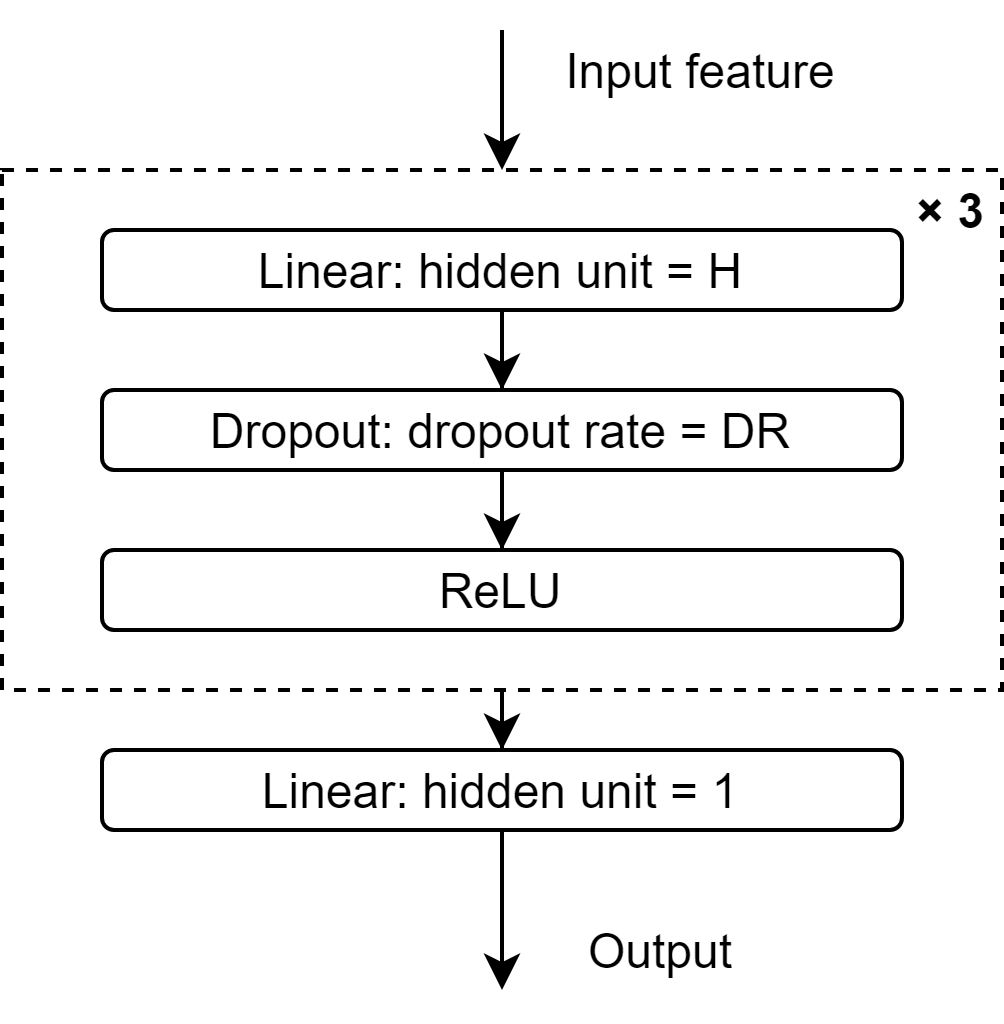}
    \caption{Model structure. Binary classifier in ORARS and GRNN share the same structure. The differences of two models are 1) the input dimension of ORARS is a concatenation of two feature vector so the input size is doubled; 2) the loss function of binary classifier in ORARS is cross-entropy, whereas the loss for GRNN training is mean square error (MSE).}
    \label{fig:model}
\end{figure}

To compare ORARS with regression method, the regression model, sORARS and ORARS are tested. The general neural-network regression (GRNN) \cite{specht1991general} is used as the baseline regression model for it had been well adopted in regression tasks
For a fairness comparison, the model structure shown in Figure.\ref{fig:model} is adopted for both GRNN and ORARS. The model contains three blocks, and each block contains a linear, a dropout \cite{hinton2012improving} and an activation layer. The activation function is ReLU \cite{nair2010rectified}. As mentioned in Section \ref{sec:reg_sorars}, the GRNN are reused to conduct the sORARS method.

The GRNN is trained with Mean Square Error (MSE) as Equation.\ref{eq:mse}
\begin{equation}
    {\mathrm {MSE}}(\bm{a},\bm{b})=\frac{1}{|\bm{a}|}\sum_{i=1}^{|\bm{a}|}(a_i-b_i)^2
    \label{eq:mse}
\end{equation}
where $\bm{a}$ and $\bm{b}$ are two vectors, $|\bm{a}|$ denotes the size of $\bm{a}$, and $a_i$ is the $i$-th element in vector $\bm{a}$. The GRNN is optimized by Adam \cite{DBLP:journals/corr/KingmaB14} algorithm.

To determine hyper-parameters for neural net, grid search proposed by Tune \cite{liaw2018tune} is adopted. The search space is shown in Table.\ref{tab:hype_grid}. Only the experiments with the minimum mean validation loss in pre-trials will be reported. The batch size and training epoch is 32 and 256, respectively.

To train ORARS, a training pair set is generated with the method described in Section \ref{sec:form}. With the training pair set, the binary classifier in ORARS is trained with Cross-entropy (CE) loss described as Equation.\ref{eq:ce_loss}
\begin{equation}
    {\mathrm {CE}}(\bm{y},\bm{p})=\frac{1}{|\bm{y}|}\sum_{i=1}^{|\bm{y}|}y_i\mathop{ln}p_i+(1-y_i)\mathop{ln}(1-p_i)
    \label{eq:ce_loss}
\end{equation}
where $y_i\in\{0,1\}$ is the $i$-th element in $\bm{y}$, representing the ground-truth of $i$-th sample, and $p_i$ is the classifier output of $i$-th sample. The Adam optimizer is applied to train the binary classifier. 

\begin{table}[h]
    \centering
    \caption{Grid search settings for GRNN. Total $|H|*|DR|*|LR|=48$ settings are conducted to find the best hyper-parameters.}
    \begin{tabular}{c|c}
        \hline
        Hyper-parameter name & Values \\ \hline
        hidden unit (H) & \{16,32,64,128\} \\
        dropout rate (DR) & \{0,0.1,0.3,0.5\}\\
        learning rate (LR) & \{0.01,0.001,0.0001\} \\
        \hline
    \end{tabular}
    \label{tab:hype_grid}
\end{table}

As proposed in \cite{su2020improving}, a weight function is applied to train the binary classifier in ORARS. Because the original weight function proposed for ordinal regression task is inappropriate for continuous variable regression tasks, A new weight function as Equation.\ref{eq:new_weight} is adopted.
\begin{equation}
    w_{ij}=\frac{|y_i-y_j|}{R}
    \label{eq:new_weight}
\end{equation}
where $y_i$ is the $i$-th sample label, and $R$ is the range value of labels. In actual training, $R$ is assigned as the maximum value minus the minimum value in the training and development sets.

The grid search described as Table.\ref{tab:hype_grid} should also be adopted in ORARS. However, with the increased dataset size, the training pair set grows exponentially and thus the time cost of pre-trials increases rapidly. Due to the computing resources limitation, a limited grid search is applied in the datasets whose size is larger than 8000 based on trails on relatively small datasets. The hyper-parameter search space in limited scenarios is \textit{$DR=0,LR\in\{0.001,0.0001\},H\in\{32,64\}$}. The search space is a subset of the space at Table.\ref{tab:hype_grid}, so if the ORARS searched in a limited space outperformed the GRNN, the advantage is clear. Similarly, only the experiments with the minimum validation loss on development set are reported. All models are trained with 8 epochs, and the batch size is 32 for the datasets whose size is smaller than 8000, or 8192 for other cases.

To explore ORARS's effectiveness and understand gains from each component, sORARS is introduced as mentioned in Section \ref{sec:reg_sorars}. By comparing GRNN with sORARS, the benefits from \textbf{Scoring with Preference} could be observed. Then, the experimental results of sORARS and ORARS will reflect pros and cons of \textbf{Binary Classifier Training} against training regressor in preference selection task.

% \paragraph{Speech Pronunciation Assessment}
% For speech pronunciation assessment task, we compared the ORARS with some traiditional machine learning methods and neural network based regression. In this task, we extract a vector which is a combanation of aGOP and cGOP for each audio, all the models are training and test on the feature.
        % Diabetes & 43 & 3.0 & 6.6 & 2\\ 
        % Pyrimidines & 74 & 0.44 & 0.9 & 27\\
        % Triazines & 186& 0.1& 0.9 & 60\\
        % Wisconsin Breast Cancer & 194& 1.0 & 125.0 & 32 \\
        % Machine CPU & 209 &6.0& 500 & 6\\
        % Auto MPG & 392 &9.0 &46.6 & 7\\
        % Boston Housing & 506 & 5.0 & 50.0& 13 \\
        % Stocks Domain & 950 & 34.0 & 62.0 & 9\\
        % Abalone & 4177 & 1.0 & 29.0 & 8\\
        % Bank Domains(1) & 8192& 0.00& 0.80&8\\
        % Bank Domains(2) & 8192& 0.00 &0.820& 32\\
        % Computer Activity(1) &8193&0.0 &99.0 & 12\\
        % Computer Activity(2) &8193&0.0 &99.0 & 21\\
        % California Housing & 20640& 14999 & 500001 &8\\
        % Census Domains(1)  & 22784& 0& 500001 &8\\
        % Census Domains(2) & 22784& 0& 500001 & 16\\ \hline

% The MAE is tested on the model with minimum loss on development set according to the grid search.
\begin{table*}[h!]
    \centering
    \caption{Dataset information and experimental results.}
    \begin{tabular}{c|c|c|c|c|c|c|c}
        \hline
        \multicolumn{5}{c|}{Dataset} & \multirow{2}{*}{GRNN} &\multirow{2}{*}{sORARS} & \multirow{2}{*}{ORARS}\\ \cline{1-5}
        Name & Size & Min & Max & Attributes & & \\ \hline
        Diabetes & 43 & 3.000 & 6.600 & 2& 0.544 & 0.538 & \textbf{0.466}\\
        Pyrimidines & 74 & 0.440 & 0.900 & 27 & 0.052 & 0.051 & \textbf{0.050}\\
        Triazines & 186& 0.100& 0.900 & 60 & 0.111 & 0.119 & \textbf{0.106}\\
        Wisconsin Breast Cancer & 194& 1.00 & 125.00 & 32 &27.29 & 30.58 & \textbf{26.59}\\        Machine CPU & 209 &6.00& 500.00 & 6 & 31.21 & \textbf{24.90}& 27.09\\
        Auto MPG & 392 &9.00 &46.60 & 7 & 2.62 & \textbf{2.22} & 2.33\\
        Boston Housing & 506 & 5.00 & 50.00& 13& 3.31 & 2.88 & \textbf{2.56}\\
        Stocks Domain & 950 & 34.00 & 62.00 & 9& 1.02 & \textbf{0.89} & 1.24 \\
        Abalone & 4177 & 1.00 & 29.00 & 8 & 1.48 & 1.54 & \textbf{1.44}\\
        Bank Domains(1) & 8192& 0.000& 0.800&8 &0.027 & 0.022 & \textbf{0.021}\\
        Bank Domains(2) & 8192& 0.000 &0.820& 32 & 0.053 & 0.054 & \textbf{0.049}\\
        Computer Activity(1) &8193&0.00 &99.00 & 12 & 8.56 & \textbf{6.86} & 7.71 \\
        Computer Activity(2) &8193 &0.00 &99.00 & 21 &6.81& 7.19 & \textbf{5.77} \\
        California Housing & 20640& 14999 & 500001 &8 & 48260 & 47170 & \textbf{41350}\\
        Census Domains(1)  & 22784& 0& 500001 &8 & 22708  & 23008 & \textbf{18466}\\ 
        Census Domains(2) & 22784& 0& 500001 & 16 & 30188 & 30458 & \textbf{23128}\\
        \hline
    \end{tabular}
    \label{tab:benchmark_result}
\end{table*}

\begin{figure*}
     \centering
     \begin{subfigure}[b]{0.33\textwidth}
         \centering
         \includegraphics[width=\textwidth]{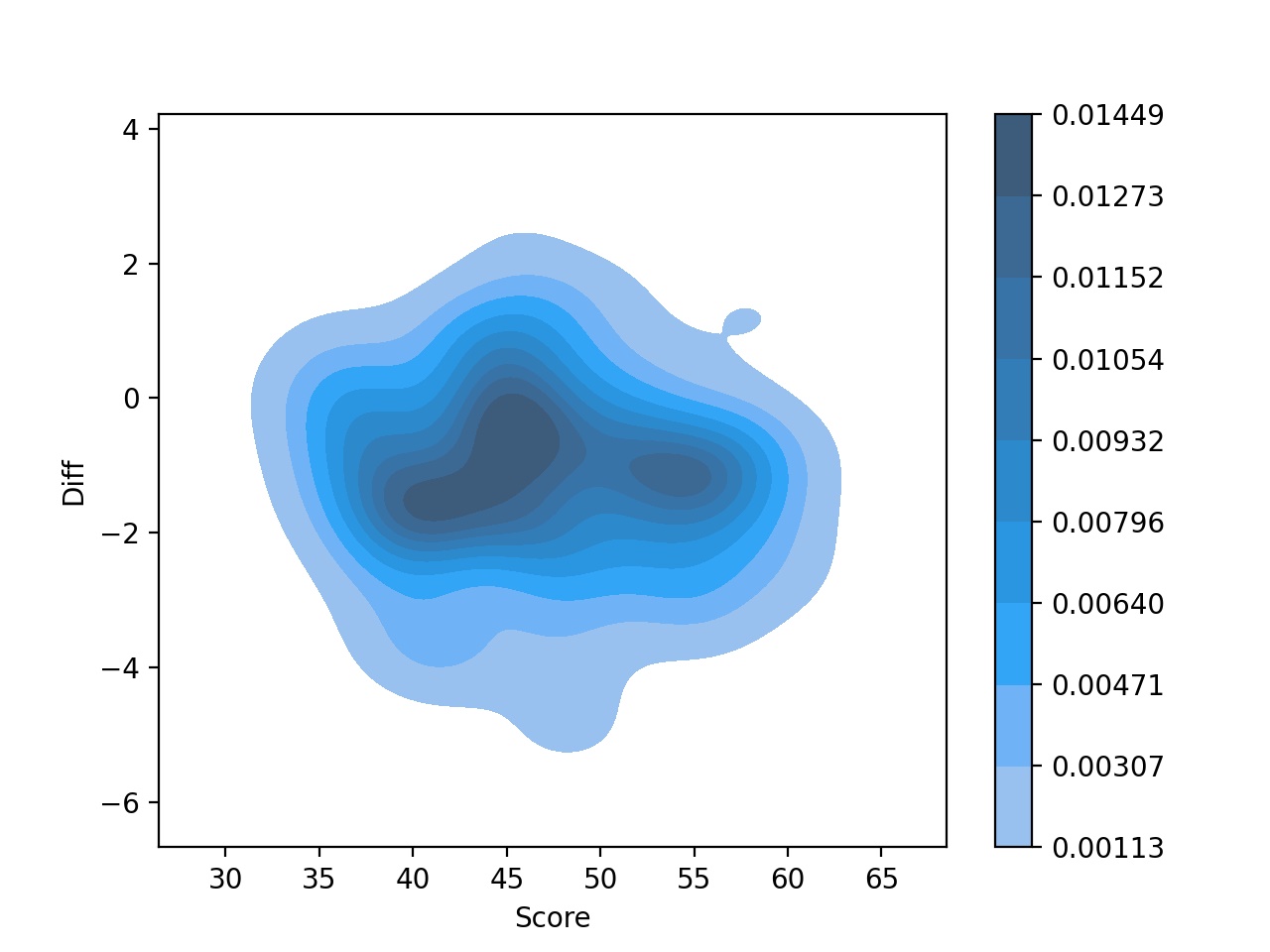}
         \caption{Stocks Domain: $1$-th fold}
         \label{fig:stock_vis}
     \end{subfigure}
     \hfill
     \begin{subfigure}[b]{0.33\textwidth}
         \centering
         \includegraphics[width=\textwidth]{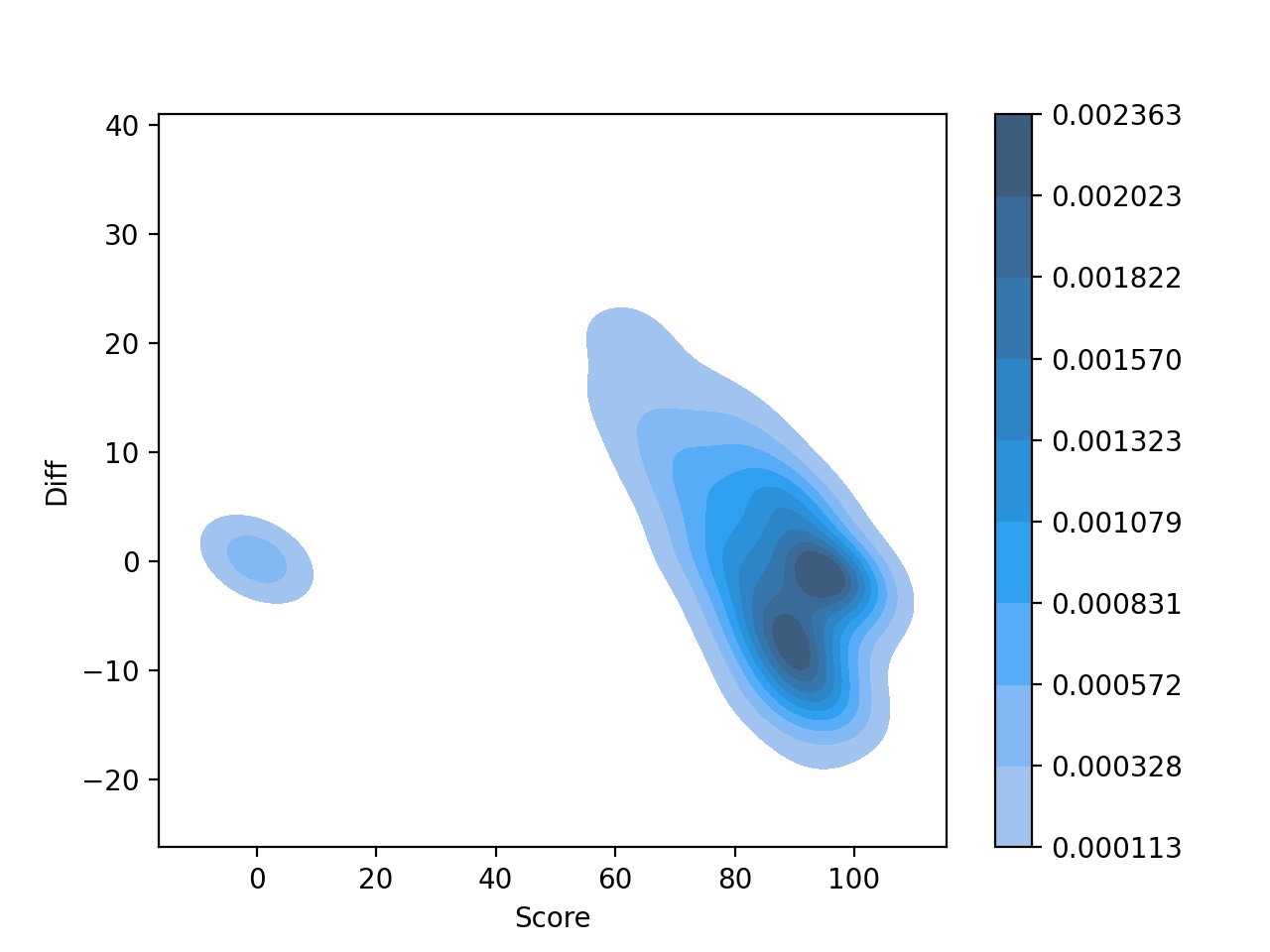}
         \caption{Computer Activity(1): $1$-th fold}
         \label{fig:bd2_vis}
     \end{subfigure}
     \hfill
     \begin{subfigure}[b]{0.33\textwidth}
         \centering
         \includegraphics[width=\textwidth]{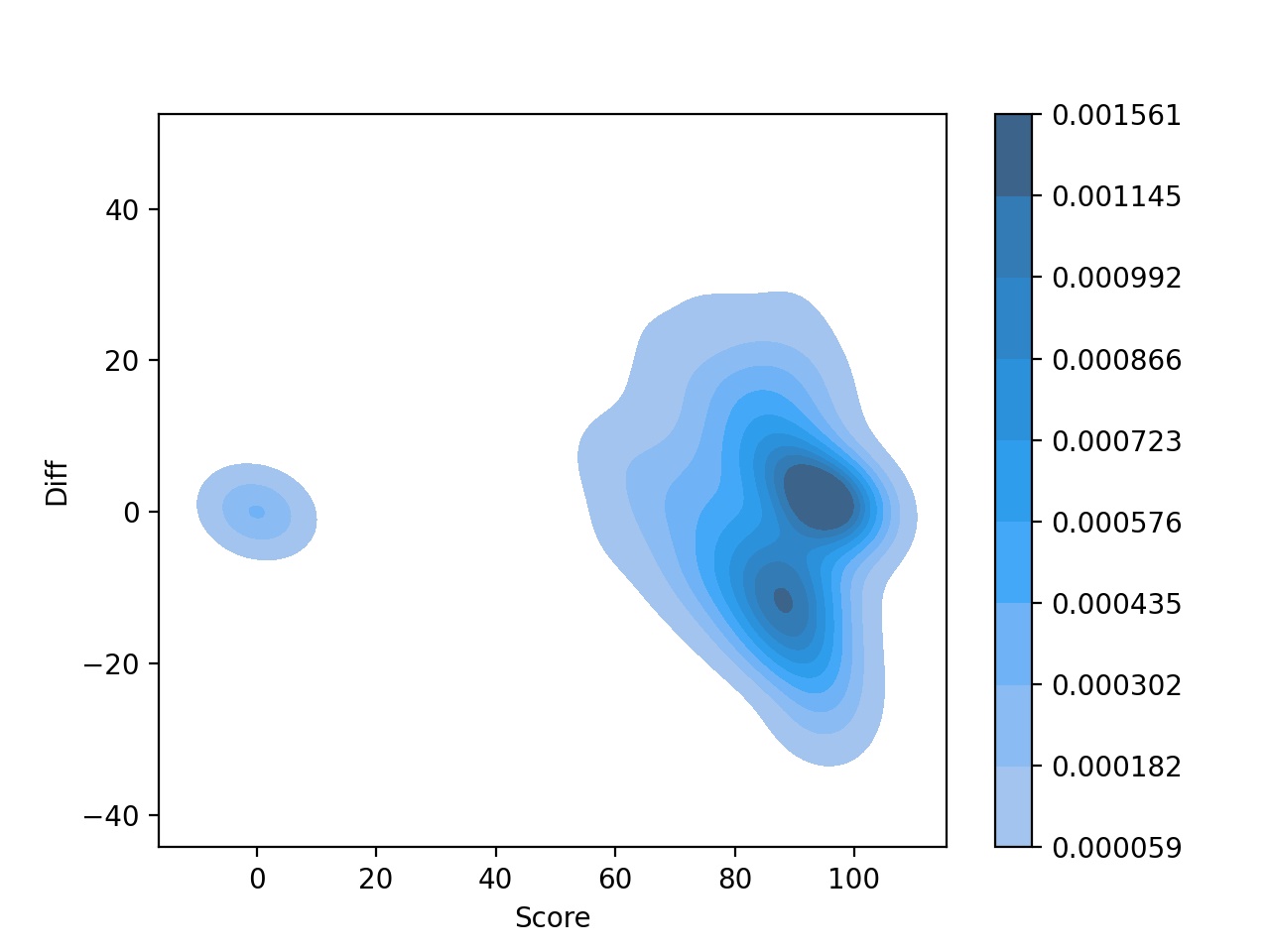}
         \caption{Computer Activity(2): $1$-th fold}
         \label{fig:ca2_vis}
     \end{subfigure}
        \caption{Visualization of the error distribution of the regressor with the best performance}
        \label{fig:vis}
\end{figure*}

\subsection{Experimental Results}

\subsubsection{Regression vs sORARS}
The experimental results are shown in Table.\ref{tab:benchmark_result}. The sORARS outperform than GRNN on 9 of 16 tasks.

Echoing the analysis in Section \ref{sec:reg_sorars}, if the regression prediction errors obey a zero-mean normal distribution, a certain improvement will obtained through sORARS in theory. For example, Figure.\ref{fig:stock_vis} shows a case where the error distribution of the best GRNN on \textbf{Stock Domain} dataset is closed to a zero-mean normal distribution. On the contrary, Figure.\ref{fig:bd2_vis} and Figure.\ref{fig:ca2_vis} and show
counterexamples where the prediction error distributions are not closed to zero-mean normal distribution. From experimental perspectives, the sORARS improve the performances on \textbf{Stock Domain} and \textbf{Computer Activity(1)} datasets but degrade the performance on \textbf{Computer Activity(2)} dataset. 

The results align with simulation analysis that when the regressor prediction error follows a zero-mean normal distribution, sORARS could improve the performance. Whereas, there is no determinate knowledge on whether the sORARS is valid or invalid under other circumstances.

\subsubsection{ORARS vs sORARS}
As results in Table.\ref{tab:benchmark_result}, by comparing sORARS with ORARS, ORARS obtains more gains on 12 of 16 datasets than sORARS. The results indicate that training a binary classifier with the generated training pairs is  better than deciding binary preference with a pre-trained regressor.

\subsubsection{Regression vs ORARS}
By comparing ORARS with GRNN, the ORARS achieves better performance on 15 of 16 datasets, showing the gains from proposed method on regression task. For counterexamples, like \textbf{Stocks Domain}, the bad performance may be caused by inappropriate network hyper-parameters or structure which lead a fail training.
% we noticed that the bad performance may be caused by inappropriate network hyper-parameters according to the observations on loss curve.

\section{Discussions}
\subsection{MOS or AB Test?}
One of the key motivations of ORARS is from media subjective evaluation, such as pronunciation scoring, text-to-speech evaluation. Riding on past data collection experiences, we observed that it is more difficult for annotators to give a sample a score of 1 to 5 than to determine which sample out of a pair is better. The previous task is called Mean Opinion Score, where an opinion score is asked to given for a simple. The second task is named as AB test, where the preference between A and B is required. Human labelers are more prone to fluctuations and deviations caused by changes in the scorer and evaluation time. But with a reference, results from AB tests are more consistent. 

Traditional regression is similar to a MOS task and ORARS is similar to an AB Test task. On the one hand, the binary comparison is simpler to determine the partial order relationships rather than a final score. On the other hand, a large number of data sample pairs are generated through the combination and pairing of the original data samples. Experimental results show that optimization on partial relationships between sample pairs can drive a better regression result, which explains that human can achieve better and consistent results when they perform AB Tests than MOS scoring.

A compromise between ORARS and regression may be the binary decomposition ordinal regression methods \cite{lin2012reduction}. These methods try to establish a classifier for each category and judge the relationship between a new sample and a given level. This approach appropriately considers ordinal information. However, ORARS turns all comparisons into a single core problem, the partial order relationship between sample pairs. According to \cite{mao2019nn} experiments, ORARS is better than binary decomposition ordinal regression methods in related tasks.

For human beings, although comparisons between sample pairs are more consistent and accurate, the number of possible combinations of sample pairs increases factorially with the number of data samples, leading to an increase in both time and expenses. But for a machine, determining the relative preference with a pair of samples can be achieved easily using a binary classifier, which does not pose a great burden on computing power. So it is a reliable solution to convert traditional regressions to a series of preference decisions.

%Optimization of binary classifier may contains a kind of regression optimization in hyperspace. The sORARS tried to train a regressor and determine which sample is better by comparing the predicted scores, on the contrary, the ORARS trys to optimize the classifier directly. With the same scoring method, the ORARS achieved better performance than sORARS and the result is better than training the regressor. It shows that optimization on partial relationships between sample pairs can drive a better regression result, which is similar to the phenomenon that human achieve better consistency when they do AB Test tasks than when they do MOS tasks.

%\subsection{sORARS vs regression}
%The standard regression is a sub-module of sORARS. To understand why the equalization could improve the score of trained regressor. A idea example is that if a regressor predict some samples with a fixed bias. By redistributing the scores of each sample, the MAE of the system can be effectively reduced. In fact, Section \ref{sec:reg_sorars} shows how this mechanism works from a probabilistic point of view under limited conditions. As Table.\ref{tab:benchmark_result} shows, the equalization worked on more than half dataset. In fact

\subsection{ORARS and Pairwise Ranking}
There are some methods for ranking problems also involving pairwise strategy or partial order relation optimization. For example, "Learning to rank" \cite{li2011short} is classical model for partial ordering, which is commonly adopted in the information retrieval field, such as \cite{burges2005learning,cao2007learning}. ORARS and the pairwise ranking have similarities and differences. Both of them take partial order information into account. But there still are some key differences. 

First, ORARS introduces the concept of anchored reference samples. With the references, a more precise ordinal information can be constructed. Second, Ranking problems aim at minimizing reversed pair numbers, whereas ORARS is to minimize the absolute prediction error. So ORARS uses training set as anchored reference samples and outputs score based on pairwise preference between test sample and anchors. Third, in ORARS a precise ranking is not the primary task, and the posterior probabilities of pairwise comparisons are accumulated to find a categorical relative ranking of the test sample in the anchor set. But for a ranking algorithm, the conflict pairwise relations are fatal, for example $A>B,B>C,A<C$, so a solution to handle conflicts is required.

\subsection{Future Works}
\subsubsection{Training Cost Optimization}
The ORARS is trained on sample pairs. For a dataset of size $|\mathbf{D}|$, the number of sample pairs is $|\mathbf{D}|^2/2$ based on self-cartesian product, leading a higher cost for training. In the inference stage, compared with the neural network-based regression, the cost is increased to $|\mathbf{D}|$ times. 

In real scenarios, even the cost of one computation is trivial, but with a supper big training set, the cost is beyond words. Two possible optimizations are: First, propose strategy to construct anchor set. Rather than directly utilize all samples in training set, some more representative samples could be picked to serve as a sampling from real data. Second, assume that the benefits of ORARS come from using a complete partial order relationship to train the model which is supported by the experiments in \cite{mao2019nn,su2020improving} and this paper. One possible way is to use parameters to learn the relationship between sample pairs. For example, for the given anchor set, anchored reference samples can be hard-coded into the model to reduce the redundant costs in the inference stage.

\subsubsection{Scoring Strategy}
\textbf{Scoring with Preference} is an important module in ORARS framework. The current scoring method is somehow a na\'ive approach. And from the statistical and experimental perspectives, there seems to be rooms for improvement. We believe that the research on a more effective way to convert partial order relationships into scores are worthwhile.

\section{Conclusion}
In this paper, we analyze the Ordinal Regression with Anchored Reference Samples (ORARS) which had been shown with a superior performance in ordinal regression tasks. By introducing a simplified ORARS (sORARS), gains from individual components in ORARS, \textbf{Binary Classifier Training} and \textbf{Scoring with Preference}, can be quantified. Experimental results on benchmark datasets show both modules are effective for regression tasks. Compared with general neural-network regression (GRNN), ORARS shows a better performance in almost all datasets. Overall speaking, in this paper, we focus on the theoretical aspects on how ORARS works and providing broad experimental results for verification. It forms a solid foundation for using ORARS as a competitive tool in general regression tasks.

\clearpage
\bibliography{example_paper}

\begin{thebibliography}{25}
\providecommand{\natexlab}[1]{#1}
\providecommand{\url}[1]{\texttt{#1}}
\expandafter\ifx\csname urlstyle\endcsname\relax
  \providecommand{\doi}[1]{doi: #1}\else
  \providecommand{\doi}{doi: \begingroup \urlstyle{rm}\Url}\fi

\bibitem[Anderson(1984)]{odd}
Anderson, J.~A.
\newblock Regression and ordered categorical variables.
\newblock \emph{Journal of the Royal Statistical Society. Series B
  (Methodological)}, 46\penalty0 (1):\penalty0 1--30, 1984.
\newblock ISSN 00359246.
\newblock URL \url{http://www.jstor.org/stable/2345457}.

\bibitem[Bender \& Grouven(1997)Bender and Grouven]{medcial1}
Bender, R. and Grouven, U.
\newblock Ordinal logistic regression in medical research.
\newblock \emph{Journal of the Royal College of Physicians of London},
  31\penalty0 (5):\penalty0 546—551, 1997.
\newblock ISSN 0035-8819.
\newblock URL \url{https://europepmc.org/articles/PMC5420958}.

\bibitem[Burges et~al.(2005)Burges, Shaked, Renshaw, Lazier, Deeds, Hamilton,
  and Hullender]{burges2005learning}
Burges, C., Shaked, T., Renshaw, E., Lazier, A., Deeds, M., Hamilton, N., and
  Hullender, G.
\newblock Learning to rank using gradient descent.
\newblock In \emph{Proceedings of the 22nd international conference on Machine
  learning}, pp.\  89--96, 2005.

\bibitem[Cao et~al.(2007)Cao, Qin, Liu, Tsai, and Li]{cao2007learning}
Cao, Z., Qin, T., Liu, T.-Y., Tsai, M.-F., and Li, H.
\newblock Learning to rank: from pairwise approach to listwise approach.
\newblock In \emph{Proceedings of the 24th international conference on Machine
  learning}, pp.\  129--136, 2007.

\bibitem[Cheng et~al.(2008)Cheng, Wang, and Pollastri]{cheng2008neural}
Cheng, J., Wang, Z., and Pollastri, G.
\newblock A neural network approach to ordinal regression.
\newblock In \emph{2008 IEEE international joint conference on neural networks
  (IEEE world congress on computational intelligence)}, pp.\  1279--1284. IEEE,
  2008.

\bibitem[Chu \& Ghahramani(2005)Chu and Ghahramani]{chu2005gaussian}
Chu, W. and Ghahramani, Z.
\newblock Gaussian processes for ordinal regression.
\newblock \emph{Journal of machine learning research}, 6\penalty0
  (Jul):\penalty0 1019--1041, 2005.

\bibitem[Diaz \& Marathe(2019)Diaz and Marathe]{Diaz_2019_CVPR}
Diaz, R. and Marathe, A.
\newblock Soft labels for ordinal regression.
\newblock In \emph{Proceedings of the IEEE/CVF Conference on Computer Vision
  and Pattern Recognition (CVPR)}, June 2019.

\bibitem[Doyle et~al.(2014)Doyle, Westman, Marquand, Mecocci, Vellas, Tsolaki,
  K{\l}oszewska, Soininen, Lovestone, Williams, et~al.]{medcial2}
Doyle, O.~M., Westman, E., Marquand, A.~F., Mecocci, P., Vellas, B., Tsolaki,
  M., K{\l}oszewska, I., Soininen, H., Lovestone, S., Williams, S.~C., et~al.
\newblock Predicting progression of alzheimer’s disease using ordinal
  regression.
\newblock \emph{PloS one}, 9\penalty0 (8):\penalty0 e105542, 2014.

\bibitem[Fernandez-Navarro et~al.(2013)Fernandez-Navarro, Campoy-Munoz,
  Hervas-Martinez, Yao, et~al.]{credit2}
Fernandez-Navarro, F., Campoy-Munoz, P., Hervas-Martinez, C., Yao, X., et~al.
\newblock Addressing the eu sovereign ratings using an ordinal regression
  approach.
\newblock \emph{IEEE transactions on cybernetics}, 43\penalty0 (6):\penalty0
  2228--2240, 2013.

\bibitem[{Gutiérrez} et~al.(2016){Gutiérrez}, {Pérez-Ortiz},
  {Sánchez-Monedero}, {Fernández-Navarro}, and {Hervás-Martínez}]{7161338}
{Gutiérrez}, P.~A., {Pérez-Ortiz}, M., {Sánchez-Monedero}, J.,
  {Fernández-Navarro}, F., and {Hervás-Martínez}, C.
\newblock Ordinal regression methods: Survey and experimental study.
\newblock \emph{IEEE Transactions on Knowledge and Data Engineering},
  28\penalty0 (1):\penalty0 127--146, 2016.
\newblock \doi{10.1109/TKDE.2015.2457911}.

\bibitem[Herbrich et~al.(1999)Herbrich, Graepel, and
  Obermayer]{herbrich1999support}
Herbrich, R., Graepel, T., and Obermayer, K.
\newblock Support vector learning for ordinal regression.
\newblock 1999.

\bibitem[Hinton et~al.(2012)Hinton, Srivastava, Krizhevsky, Sutskever, and
  Salakhutdinov]{hinton2012improving}
Hinton, G.~E., Srivastava, N., Krizhevsky, A., Sutskever, I., and
  Salakhutdinov, R.~R.
\newblock Improving neural networks by preventing co-adaptation of feature
  detectors.
\newblock \emph{arXiv preprint arXiv:1207.0580}, 2012.

\bibitem[Kingma \& Ba(2015)Kingma and Ba]{DBLP:journals/corr/KingmaB14}
Kingma, D.~P. and Ba, J.
\newblock Adam: {A} method for stochastic optimization.
\newblock In Bengio, Y. and LeCun, Y. (eds.), \emph{3rd International
  Conference on Learning Representations, {ICLR} 2015, San Diego, CA, USA, May
  7-9, 2015, Conference Track Proceedings}, 2015.
\newblock URL \url{http://arxiv.org/abs/1412.6980}.

\bibitem[Kwon et~al.(1997)Kwon, Han, and Lee]{credit1}
Kwon, Y.~S., Han, I., and Lee, K.~C.
\newblock Ordinal pairwise partitioning (opp) approach to neural networks
  training in bond rating.
\newblock \emph{Intelligent Systems in Accounting, Finance \& Management},
  6\penalty0 (1):\penalty0 23--40, 1997.

\bibitem[Li(2011)]{li2011short}
Li, H.
\newblock A short introduction to learning to rank.
\newblock \emph{IEICE TRANSACTIONS on Information and Systems}, 94\penalty0
  (10):\penalty0 1854--1862, 2011.

\bibitem[Liaw et~al.(2018)Liaw, Liang, Nishihara, Moritz, Gonzalez, and
  Stoica]{liaw2018tune}
Liaw, R., Liang, E., Nishihara, R., Moritz, P., Gonzalez, J.~E., and Stoica, I.
\newblock Tune: A research platform for distributed model selection and
  training.
\newblock \emph{arXiv preprint arXiv:1807.05118}, 2018.

\bibitem[Lin \& Li(2012)Lin and Li]{lin2012reduction}
Lin, H.-T. and Li, L.
\newblock Reduction from cost-sensitive ordinal ranking to weighted binary
  classification.
\newblock \emph{Neural Computation}, 24\penalty0 (5):\penalty0 1329--1367,
  2012.

\bibitem[Mao et~al.(2019)Mao, Wu, Jiang, Liu, and Soong]{mao2019nn}
Mao, S., Wu, Z., Jiang, J., Liu, P., and Soong, F.~K.
\newblock Nn-based ordinal regression for assessing fluency of esl speech.
\newblock In \emph{ICASSP 2019-2019 IEEE International Conference on Acoustics,
  Speech and Signal Processing (ICASSP)}, pp.\  7420--7424. IEEE, 2019.

\bibitem[McCullagh(1980)]{mccullagh1980regression}
McCullagh, P.
\newblock Regression models for ordinal data.
\newblock \emph{Journal of the Royal Statistical Society: Series B
  (Methodological)}, 42\penalty0 (2):\penalty0 109--127, 1980.

\bibitem[Nair \& Hinton(2010)Nair and Hinton]{nair2010rectified}
Nair, V. and Hinton, G.~E.
\newblock Rectified linear units improve restricted boltzmann machines.
\newblock In \emph{Icml}, 2010.

\bibitem[Perez-Orti{\'z} et~al.(2013)Perez-Orti{\'z}, Guti{\'e}rrez, and
  Herv{\'a}s-Mart{\'\i}nez]{perez2013projection}
Perez-Orti{\'z}, M., Guti{\'e}rrez, P.~A., and Herv{\'a}s-Mart{\'\i}nez, C.
\newblock Projection-based ensemble learning for ordinal regression.
\newblock \emph{IEEE transactions on cybernetics}, 44\penalty0 (5):\penalty0
  681--694, 2013.

\bibitem[Specht et~al.(1991)]{specht1991general}
Specht, D.~F. et~al.
\newblock A general regression neural network.
\newblock \emph{IEEE transactions on neural networks}, 2\penalty0 (6):\penalty0
  568--576, 1991.

\bibitem[Su et~al.(2020)Su, Mao, Soong, Xia, Tien, and Wu]{su2020improving}
Su, B., Mao, S., Soong, F., Xia, Y., Tien, J., and Wu, Z.
\newblock Improving pronunciation assessment via ordinal regression with
  anchored reference samples.
\newblock \emph{arXiv preprint arXiv:2010.13339}, 2020.

\bibitem[Torra et~al.(2006)Torra, Domingo-Ferrer, Mateo-Sanz, and
  Ng]{torra2006regression}
Torra, V., Domingo-Ferrer, J., Mateo-Sanz, J.~M., and Ng, M.
\newblock Regression for ordinal variables without underlying continuous
  variables.
\newblock \emph{Information Sciences}, 176\penalty0 (4):\penalty0 465--474,
  2006.

\bibitem[Xia et~al.(2007)Xia, Zhou, Yang, and Zhang]{xia2007ordinal}
Xia, F., Zhou, L., Yang, Y., and Zhang, W.
\newblock Ordinal regression as multiclass classification.
\newblock \emph{International Journal of Intelligent Control and Systems},
  12\penalty0 (3):\penalty0 230--236, 2007.

\end{thebibliography}
\bibliographystyle{icml2021}

\appendix
% \section{Do \emph{not} have an appendix here}

% \textbf{\emph{Do not put content after the references.}}
% %
% Put anything that you might normally include after the references in a separate
% supplementary file.

% We recommend that you build supplementary material in a separate document.
% If you must create one PDF and cut it up, please be careful to use a tool that
% doesn't alter the margins, and that doesn't aggressively rewrite the PDF file.
% pdftk usually works fine. 

% \textbf{Please do not use Apple's preview to cut off supplementary material.} In
% previous years it has altered margins, and created headaches at the camera-ready
% stage. 
%%%%%%%%%%%%%%%%%%%%%%%%%%%%%%%%%%%%%%%%%%%%%%%%%%%%%%%%%%%%%%%%%%%%%%%%%%%%%%%
%%%%%%%%%%%%%%%%%%%%%%%%%%%%%%%%%%%%%%%%%%%%%%%%%%%%%%%%%%%%%%%%%%%%%%%%%%%%%%%

\end{document}